\documentclass[10pt,twocolumn,letterpaper]{article}

\usepackage[pagenumbers]{cvpr} 

\usepackage{graphicx}
\usepackage{amsmath}
\usepackage{amssymb}
\usepackage{booktabs}
\usepackage{times}
\usepackage{epsfig}
\usepackage{multirow}
\usepackage{arydshln}
\usepackage{enumitem}

\usepackage[pagebackref,breaklinks,colorlinks]{hyperref}

\usepackage[capitalize]{cleveref}
\crefname{section}{Sec.}{Secs.}
\Crefname{section}{Section}{Sections}
\Crefname{table}{Table}{Tables}
\crefname{table}{Tab.}{Tabs.}

\def\cvprPaperID{*****} 
\def\confName{CVPR}
\def\confYear{2023}

\begin{document}

\title{Open-Vocabulary Object Detection using Pseudo Caption Labels}

\author{
Han-Cheol Cho\thanks{Corresponding author.} \space\space\space\space\space Won Young Jhoo \space\space\space\space\space Wooyoung Kang \space\space\space\space\space Byungseok Roh\\
Multi-modal Understanding Team, KakaoBrain\\
{\tt\small \{simon.cho, iji.young, edwin.kang, peter.roh\}@kakaobrain.com}
}

\maketitle

\begin{abstract}
Recent open-vocabulary detection methods aim to detect novel objects by distilling knowledge from vision-language models (VLMs) trained on a vast amount of image-text pairs.
To improve the effectiveness of these methods, researchers have utilized datasets with a large vocabulary that contains a large number of object classes, under the assumption that such data will enable models to extract comprehensive knowledge on the relationships between various objects and better generalize to unseen object classes.
In this study, we argue that more fine-grained labels are necessary to extract richer knowledge about novel objects, including object attributes and relationships, in addition to their names.
To address this challenge, we propose a simple and effective method named Pseudo Caption Labeling (PCL), which utilizes an image captioning model to generate captions that describe object instances from diverse perspectives.
The resulting pseudo caption labels offer dense samples for knowledge distillation.
On the LVIS benchmark, our best model trained on the de-duplicated \emph{VisualGenome} dataset achieves an AP of 34.5 and an APr of 30.6, comparable to the state-of-the-art performance.
PCL's simplicity and flexibility are other notable features, as it is a straightforward pre-processing technique that can be used with any image captioning model without imposing any restrictions on model architecture or training process.
Code is available at: \url{TBA}.
\end{abstract}

\section{Introduction}
\label{sec:introduction}

\begin{figure}
    \centering
    \includegraphics[width=0.45\textwidth]{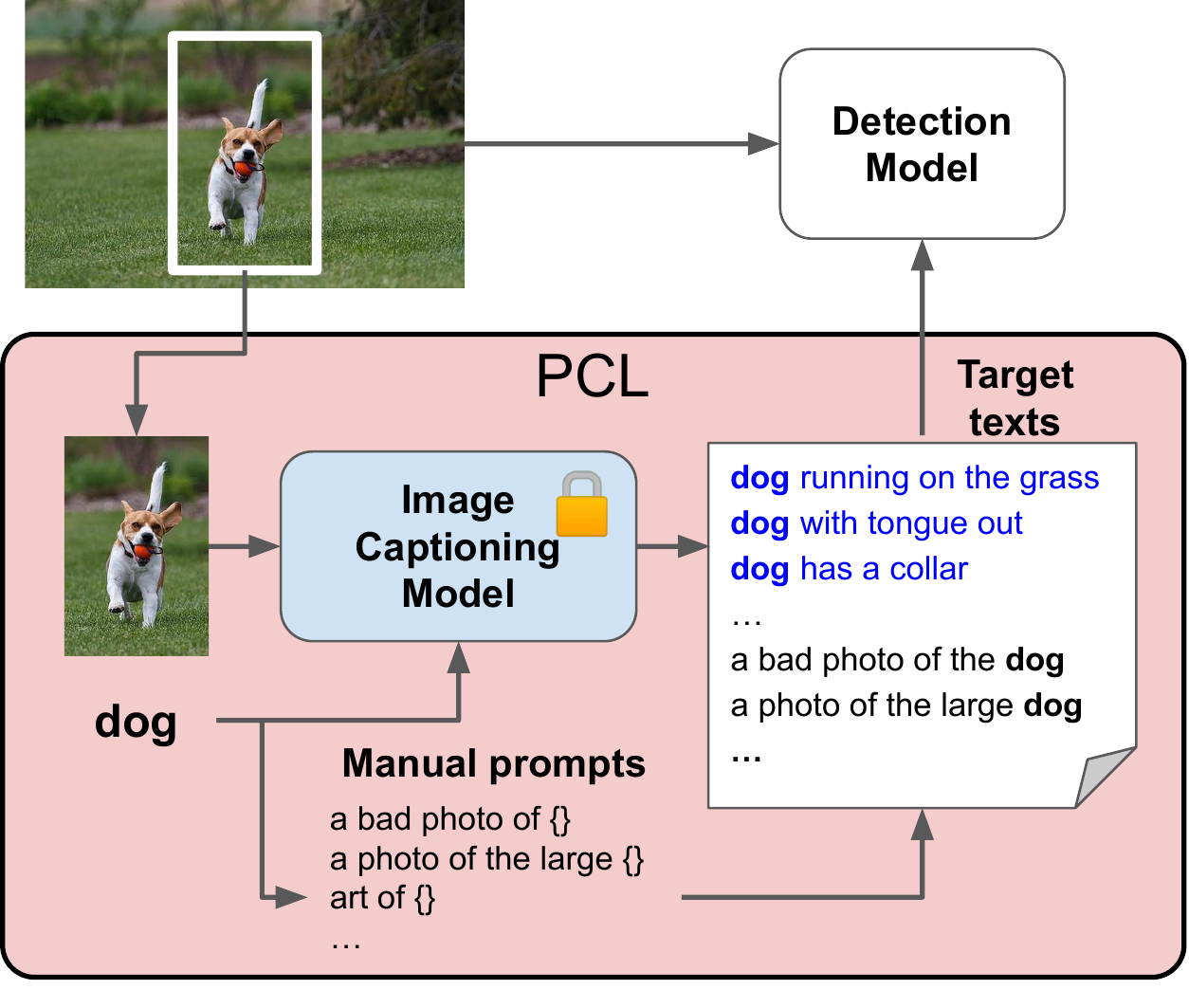}
    \caption{System overview. PCL generates pseudo caption labels (shown in blue) by employing an image captioning model. These labels provide detailed information about the main object and its surroundings compared to the manual prompt labels.}
    \label{fig:system_overview}
\end{figure}

\begin{figure*}[t!]
    \centering
    \includegraphics[width=1.0\textwidth]{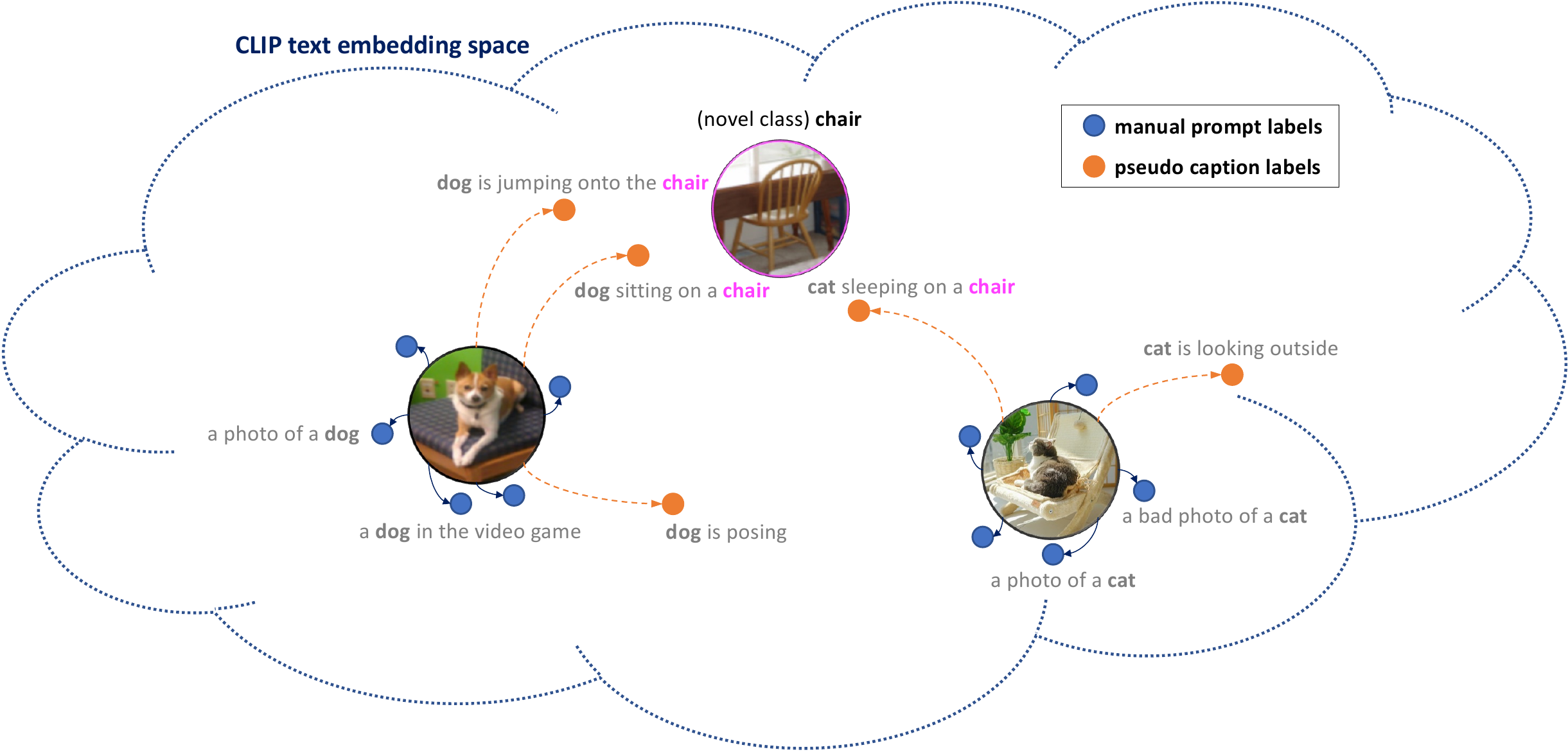}
    \caption{An illustration of the distillation points for two object instances, a cat and a dog, in the CLIP text latent space. The blue points represent manual prompt labels, while the orange ones indicate pseudo caption labels. Some of the pseudo caption labels are referring to \textbf{chair}, which is a novel object class. It is expected that their embeddings are located near the chair class embedding.}
    \label{fig:latent_space}
\end{figure*}

Open-vocabulary detection (OVD) is a challenging task that aims to identify novel objects not present in the training data.
Recent studies~\cite{ZSD2018,ViLD2021,OWLViT2022,CentricOVD2022,RegionCLIP2022} have shown the potential of leveraging vision-language models (VLMs) trained on a vast amount of image-text pairs to achieve this goal.
Many of them use datasets with a large vocabulary to better distill knowledge from VLMs.
Previous research~\cite{GLIP2022,OWLViT2022} has utilized multiple datasets such as Object365~\cite{Object365_2019}, OpenImages~\cite{OpenImages}, and VisualGenome~\cite{VisualGenome2017} to acquire such data.
Alternatively, another approach~\cite{PBOvd2021,VLDet2023,RegionCLIP2022} acquires annotated data through a weakly-supervised manner.

Utilizing a large vocabulary dataset is necessary, but may not be sufficient to extract comprehensive knowledge about novel objects.
VLMs are trained on image-text pairs~\cite{ALIGN2021,CLIP2021}, where the accompanying text provides diverse information such as object names, their attributes and relationships, and the surrounding environments in which they exist.
As a result, object names or concepts in VLM's latent space are associated with many different types of information.
If we can utilize such diverse information, it will be possible to extract rich knowledge about novel objects from VLMs.

To address this problem, we propose a simple and effective method called \textit{Pseudo Caption Labeling} (PCL).
PCL is a pre-processing technique that utilizes an image-captioning model to transform the class name of an object instance into descriptive captions, as shown in Figure~\ref{fig:system_overview}.
More specifically, we generate these pseudo caption labels by feeding a cropped object image with a 20\% margin into the captioning model.
Additionally, the object name is used as the output prefix to ensure that the generated caption labels explicitly mention the target object.

Figure~\ref{fig:latent_space} illustrates how pseudo caption labels enable a model to better distill knowledge about novel objects.
The figure shows three object instances, where two of them are from known classes, cat and dog, and the other is from unseen category, chair.
Although the training dataset does not have object annotations for the chair category, the pseudo caption labels for the cat and dog class objects may describe the novel cateogry object chair such as ``dog sitting on a chair'' and ``cat sleeping on a chair''.
These pseudo caption labels representing the relationship between two objects are likely to be located between the class embeddings of the objects in the VLM's latent space, allowing the model extract knowledge closer to the novel object class more effectively.

Our best model, which was trained on the de-duplicated VisualGenome (VG) dataset, achieved an AP of 34.5 and an APr of 30.6 on the LVIS benchmark, which are comparable to the previous state-of-the-art methods that used multiple datasets.
Ablation experiments revealed that various design choices we made, such as using multiple captions per object instance and using class name as output prefix, were crucial for achieving such performance.
Another key feature is its simplicity and flexibility, as it is a straightforward pre-processing technique that can be used with any image-captioning model without imposing any restrictions on model architecture or training process.
Given these advantages, we believe that PCL will serve as a building block for a robust baseline in future OVD research.

\section{Related Work}
\label{sec:related_work}

\subsection{Vision-Language Pre-training}

Modality-specific pre-training is a widely used approach that improves performance on downstream tasks.
For instance, in computer vision, pre-trained models on the ImageNet classification task~\cite{ImageNet2009} are commonly used.
In natural language processing area, models are often pre-trained on a large-scale unlabeled text data such as Wikipeidia, BookCorpus~\cite{BookCorpus2015}, and Colossal Clean Crawled Corpus~\cite{C4_2020}.

Vision-language pre-training (VLP)~\cite{SimVLM2022,ALIGN2021,VILT2021,CLIP2021} takes this approach one step further by learning a shared latent space where two different modalities are aligned together.
This shared latent space allows for zero-shot generalization, a unique characteristic of VLP.
For example, CLIP \cite{CLIP2021} achieved impressive results on zero-shot image classification on various benchmark datasets.
Most OVD methods, including ours, also rely on the use of VLMs.

\subsection{Open-Vocabulary Object Detection}

Open-vocabulary detection methods can be classified into three categories.
The first category consists of methods that utilize image-text pairs.
For example, RegionCLIP~\cite{RegionCLIP2022} trains a region-level CLIP model on region-concept pairs created from an image-caption dataset.
When fine-tuned on a detection dataset, it outperforms the one using the original CLIP model.
VLDet~\cite{VLDet2023} trains a model on a detection dataset and an image-text pair dataset simultaneously.
For an input consisting of an image-text pair, it extracts all nouns from the caption, matches them with region proposals using a bipartite matching algorithm, and uses them as ground-truth labels.

The methods in the second category leverage both the visual and linguistic latent spaces of VLMs. 
For instance, ViLD~\cite{ViLD2021} uses an additional loss to minimize the L1 distance between the predicted class embeddings of region proposals and the corresponding CLIP image embeddings.
Object-centric OVD~\cite{CentricOVD2022} introduces inter-embedding relationship loss to ensure that the relationship between two region embeddings is consistent with that of CLIP image embeddings.

The methods in the last category employ the image encoder of VLMs directly as the detector's backbone.
OWL-ViT~\cite{OWLViT2022} constructs a detection model by attaching MLP heads on top of the CLIP image encoder.
F-VLM~\cite{FVLM2023} uses a similar approach to OWL-ViT, but does not train the backbone during training.
These methods can effectively utilize the knowledge in VLM's visual latent space and typically outperform other approaches.

Compared to these approaches, PCL is an extension of the most fundamental approach, focusing solely on extracting knowledge from VLM's linguistic latent space by using only a detection dataset.
We believe that combining our approach with previous methods could lead to enhanced performance, but we'll leave this for future studies.

\section{Method}
\label{sec:method}

In this section, we explain three main components of the proposed methods: an image captioning model, a pseudo caption label generation method, and an open-vocabulary detection model.

\subsection{Image Captioning Model}
\label{sec:method:captioner}

Given an image $I$ and a corresponding caption $c$ consisting of $T$ words ($w_1, ..., w_T)$, an image captioning model is typically trained by minimizing the negative log-likelihood, as follows: 
\begin{equation}
     \mathcal{L} ={-\log p(c|I)} = \sum_{t=0}^{T}{-\log p(w_{t+1}|w_{\le t}, I)},
    \label{eq:log-likelihood}
\end{equation}
where the BOS and EOS tokens ($w_0$ and $w_{T+1}$) represent the start and end of a sentence, respectively. 
Following the recent image captioning works~\cite{Blip2022,Git2022,VinVL2021}, we train the model on a combination of large-scale captioning datasets consisting of Conceptual Captions 3M (CC3M)~\cite{CC3M2018}, Conceptual Captions 12M (CC12M)~\cite{CC12M2021}, MSCOCO~\cite{MSCOCO2014}, and Visual Genome (VG)~\cite{VisualGenome2017}.
For VG, we use all three types of annotations that describe regions, attributes, and relationships.

While large and diverse datasets can enhance the visual-language understanding of a captioning model, it can also lead to the generation of captions that include irrelevant details, such as background objects and scenery, due to the influence of caption styles in CC3M, CC12M, and MSCOCO.
To address this issue, we introduced a style conditioning mechanism that controls output caption styles.
Formally, Equation~\ref{eq:log-likelihood} can be re-formulated as follows:
\begin{equation}
     \mathcal{L} ={-\log p(c|I, z)} = \sum_{t=0}^{T}{-\log p(w_{t+1}|w_{\le t}, I, z)},
    \label{eq:cond-log-likelihood}
\end{equation}
where the variable $z$ represents the style conditioning vector that corresponds to the training dataset identities.
To incorporate the style vector, we concatenate the style vector $z$ to the visual feature vector and feed the combined vector to the caption decoder as a prefix context.

By using style conditioning, our conditional captioning model can reliably produce captions that accurately describe images in a desired format.
We have chosen the VG-region style as the style condition for our experiments, because it results in generated captions that effectively describe the primary objects, their attributes and their relationships with the surrounding environment, all in a brief and succinct manner.
More details about our conditional captioning model are explained in Section~\ref{exp:impl_details_cap}.

\subsection{Pseudo Caption Label Generation}
\label{sec:method:pcl}

To generate pseudo caption labels for each object in the training dataset, we use the captioning model explained earlier.
This is done through a 4-step process:
\begin{enumerate}
  \item Crop an object image
  \item Feed the cropped image to the captioning model
  \item Generate captions for the object, utilizing the object name as the output prefix
  \item Collect K distinct captions for the object
\end{enumerate}

Pseudo caption labels provide a model much diverse information for knowledge distillation compared to manual prompt labels.
Nevertheless, during testing, we still employ manual prompts labels, which can lead to a decline in performance due to the discrepancy of target labels between the training and testing phases.
To mitigate this issue, we also use a small portion of manual prompt labels for training.
We show its impact in the ablation experiments in Section \ref{exp:ablations}.

\subsection{Detection Model}
\label{sec:method:detector}

DETR~\cite{DETR2020} is a Transformer-based object detection model that solves object detection as a set prediction problem, utilizing a bipartite matching algorithm.
Our detection model is based on Deformable-DETR~\cite{DeformDETR2021}, which improves the convergence speed of DETR using a deformable attention mechanism.
For open-vocabulary detection, we replaced the model's last classification layer with pre-trained CLIP text embeddings of target classes, following the previous work~\cite{ZSD2018}.

The model operates similarly to the original Deformable DETR, with the decoder producing class embeddings for object queries.
To compute the class probability of the $j$-th target class $c_j$ given the $i$-th object query $o_i$, we use the following equation:
\begin{equation}
    p(c_j|o_i) = s(\hat{e}_{i}, e_{j}) = \sigma\left(a\frac{\hat{e}_{i} \cdot e_{j}}{|\hat{e}_{i}||e_{j}|} + b\right).
    \label{eq:embedding_similarity}
\end{equation}
Here, $s$ denotes the score function, $\hat{e}_i$ represents the predicted class embedding of the object query $o_i$, and $e_{j}$ is the target class embedding generated by applying the CLIP text encoder to one of pseudo caption labels (or manual prompt labels) for the target class $c_j$.
$a$ is the logit scale, $b$ is the logit shift, and $\sigma$ is the sigmoid function.

Then, we calculate the classification loss using Equation~\ref{eq:classification_loss}.
In this equation, $N$ is the number of object queries, $M$ is the number of target classes and sampled negative classes, and $y_{i, j}$ equals 1 if the object query $o_i$ and the target class $c_j$ form a positive pair obtained via a bipartite matching, and 0 otherwise.
\begin{equation}
    \mathcal{L}_{class} = \frac{1}{N}\sum^{N}_{i=1}\sum^{M}_{j=1}{BCE(p(c_j|o_i), y_{i,j})}
    \label{eq:classification_loss}
\end{equation}

All other components are identical to those in the original Deformable DETR.
In addition, it is noteworthy that our approach can be applied to other object detection architectures, including Faster-RCNN~\cite{RCNN2015}.

\section{Experiments}
\label{sec:experiments}

In this section, we evaluate the proposed method on the LVIS benchmark, demonstrate the impact of our design choices through ablation experiments, and show the PCL model's ability to comprehend complex queries with qualitative analysis.

\subsection{Implementation Details}
\label{exp:impl_details}

\subsubsection{Captioning Model}
\label{exp:impl_details_cap}

Our style-conditional captioning model consists of two main modules: an image encoder and a caption decoder. We initialize the image encoder and the caption decoder using pre-trained CLIP ViT-L/14~\cite{CLIP2021} and GPT2~\cite{gpt2}, respectively. The model trained for 10 epochs using widely used captioning datasets, including CC3M~\cite{CC3M2018}, CC12M~\cite{CC12M2021}, and MSCOCO~\cite{MSCOCO2014}. Additionally, we further utilize region, attribute, and relation annotations in VG~\cite{VisualGenome2017} dataset. During the training of our model, we use AdamW optimizer~\cite{adamw} with a linear warm-up strategy for the first epoch followed by a learning rate decaying with a cosine schedule. We set a base learning rate as 0.0016 with a total batch size of 2048.

For style conditioning, we categorize the datasets into four styles: a) \textit{Scene description}, b) \textit{Attribute description}, c) \textit{Relation description}, and d) \textit{Region description}. We assign CC3M, CC12M, and MSCOCO to the style a) and the attribute, relation, and region annotations of VG are assigned to the styles b), c), and d), respectively. 
During the training phase, given an image-text pair, we firstly assign a style to the pair according to its dataset identity (\eg, \textit{"Relation description"} if the pair comes from relation annotation of VG). Then, we concatenate the CLS token of visual features from the image encoder and the parsed word embeddings of style keywords. After the concatenation, the resulting features are utilized as a prefix sequence of our caption decoder model, \ie, GPT2. 
At the inference time, we use \textit{Region description} style condition since it is the most favorable to the OVD task.

\subsubsection{Detection Model}
\label{exp:impl_details}

Our detection model is a modified Deformable DETR described in Section \ref{sec:method:detector}.
The following settings apply to all models and experiments, unless otherwise specified.
To handle a large vocabulary dataset, we increase the number of object queries from 100 to 300.
We also activate iterative box-refinement and two-stage prediction options.
The Swin-T model serves as the backbone.
We train models for 80K steps with an effective batch size of 128 by accumulating gradients for two batches (64$\times$2).
It is approximately 100 epochs for the LVIS base and de-duplicated VG datasets.
The initial learning rate is $2\times10^{-4}$ and decreases to $2\times10^{-5}$ after 65K steps.
Additionally, we keep the logit scale and logit shift fixed at 25 and -0.3, respectively, as our preliminary experiments have shown that they not only reduce training time but also does not hurt final performance~\cite{ItalianCLIP2021}.
For the baseline models, we use RFS sampling~\cite{RFS_2018} with the threshold 0.001, while no sampling methods are used for PCL models.

Baseline models uses 80 labels per object, which are generated from the prompt templates introduced in the previous work~\cite{CLIP2021}.
PCL models use 20 labels per object, consisting of 80\% (16) pseudo caption labels and 20\% (4) manual prompt labels.
Manual prompt labels are created by sampling from the seven best prompt templates~\cite{CLIP2021} to better match their label patterns with those used for testing.
During training, PCL models randomly sample one of 20 labels as the target label, following a commonly used method in previous OVD works.
However, for baseline models, we use the mean embedding of the 80 labels, as it demonstrated slightly higher performance in our preliminary experiments.

Choosing an appropriate negative class sampling method is crucial when training models on a large vocabulary dataset.
For the baseline model trained on the LVIS base, we samples 50 negative classes based on the square-root class frequency in the training data~\cite{CenterNetV2_2021}.
For all other models, we use target embeddings collected from all GPUs and a memory bank, which keeps a fixed amount of unique target embeddings during training.
We set the memory bank size to 200 and update 10 oldest entries with new ones every training step.
This approach is not dependent on the number of target classes in the training data, making it much easier to use in conjunction with PCL.
The choice of a negative class sampling method is determined by the model's performance in preliminary experiments.

Our models are trained on two datasets: the LVIS base and the de-duplicated VG datasets.
LVIS~\cite{LVIS2019} is a large vocabulary dataset, consisting of 1,203 object categories with three class types, frequent, common and rare.
The LVIS base is the official training split for OVD evaluation comprised of only frequent and common type objects.
VG~\cite{VisualGenome2017} is a manually annotated dataset with the largest vocabulary size, including over 80,000 categories.
The de-duplicated VG is its subset built by removing the LVIS validation images and the rare category annotations~\cite{OWLViT2022}.
We selected VG because PCL can take advantage of a larger vocabulary dataset to provide the model with more diverse object-dependent contextual information, leading to significant enhancements in OVD performance.
To encode target labels, we use the CLIP text encoder of the ViT-L/14 model.

Our evaluation metric is based on box-AP and box-APr, as our detector does not generate segmentation masks.
For evaluation, we use the seven best prompt templates~\cite{CLIP2021} with all category names of the LVIS dataset.
We ensemble the class prediction results over the prompts like previous work~\cite{OWLViT2022}.

\begin{table}[t!]
\begin{center}
\begin{tabular}{c c c c c c c} 
 \hline
 Training Data                & Method          & APr & AP \\
 \hline\hline
 \multirow{2}{*}{LVIS base}   & Baseline        & 18.7 & 34.5 \\ 
                              & PCL (ours)      & 20.1 (+1.4) & 33.5 \\
 \hline
 \multirow{2}{*}{VG dedup}    & Baseline        & 21.1 & 24.3 \\ 
                              & PCL (ours)      & 25.7 (+4.6) & 29.5 \\
\hdashline
 \multirow{2}{*}{VG all}      & Baseline\dag    & 25.3 & 26.6 \\
                              & PCL (ours)\dag  & 28.5 & 30.8 \\
 \hline
\end{tabular}
\end{center}
\caption{Evaluation results on the LVIS benchmark. PCL consistently improves the novel object detection performance (APr). Notably, the improvement is more significant when the method is applied to the de-duplicated VG dataset, which contains objects over 80K categories. Furthermore, we provide an upper-bound performance, denoted with the \dag symbol, without de-duplication.}
\label{table:main_result_1}
\end{table}

\begin{table*}[t!]
\begin{center}
\scalebox{0.95}{
\begin{tabular}{c c c c c c c c c} 
 \hline
 Method & Backbone & Params. & Detector & Detection Data & Image Data & KD Source & APr & AP \\
 \hline\hline
 GLIP       & Swin-T & 28M & GLIP        & O365, GoldG & Cap4M   & -                      & 10.1 & 17.2 \\
 ViLD-text  & RN50   & 26M & Mask-RCNN   & LVIS base   & -       & CLIP$_{text}$ ViT-B/32 & 10.6 & 27.9 \\
 ViLD-ens.  & RN50   & 26M & Mask-RCNN   & LVIS base   & -       & CLIP ViT-B/32          & 16.7 & 27.8 \\
 RegionCLIP & RN50   & 26M & Faster-RCNN & LVIS base   & CC3M    & CLIP ViT-B/32          & 17.1 & 28.2 \\
 VLDet      & RN50   & 26M & CenterNetV2 & LVIS base   & CC3M    & CLIP$_{text}$ RN50     & 22.9 & 33.4 \\
 PCL (ours) & Swin-T & 28M & Def.DETR    & VG dedup    & -       & CLIP$_{text}$ ViT-L/14 & \underline{\textbf{25.7}} & 29.5 \\
 \hline
 OWL-ViT    & ViT-B/16 & 86M & MLP         & O365, VG dedup & -    & CLIP ViT-B/16          & 20.6 & 27.2 \\
 ViLD-ens.  & EN-B7    & 67M & Mask-RCNN   & LVIS base      & -    & CLIP ViT-L/14          & 22.0 & 32.4 \\
 ViLD-ens.  & EN-B7    & 67M & Mask-RCNN   & LVIS base      & -    & ALIGN                  & 27.0 & 31.8 \\
 VLDet      & Swin-B   & 88M & CenterNetV2 & LVIS base      & CC3M & CLIP$_{text}$ RN50     & \textbf{29.6} & 42.7 \\
 PCL (ours) & Swin-B   & 88M & Def.DETR    & VG dedup       & -    & CLIP$_{text}$ ViT-L/14 & \underline{29.1} & 32.9 \\
 \hline
 GLIP       & Swin-L   & 197M & GLIP      & OI, O365, ...  & CC12M, SBU  & -            & 17.1 & 26.9 \\
 OWL-ViT    & ViT-L/14 & 303M & MLP       & O365, VG dedup & - & CLIP ViT-L/14          & \textbf{31.2} & 34.6 \\
 PCL (ours) & Swin-L   & 197M & Def.DETR  & LVIS base      & - & CLIP$_{text}$ ViT-L/14 & 24.7 & 38.7 \\
 PCL (ours) & Swin-L   & 197M & Def.DETR  & VG dedup       & - & CLIP$_{text}$ ViT-L/14 & \underline{30.6} & 34.5 \\
 \hline
\end{tabular}
}
\end{center}
\caption{Comparison with other methods. We categorize the methods into three groups based on backbone size. The first group includes ResNet50 (RN50) and Swin-T, the second group includes EfficientNet-B7 (EN-B7), ViT-B/16, and Swin-B, and the third group includes Swin-L and ViT-L/14. For all detection datasets, we used their de-duplicated versions. GLIP with Swin-L uses OI, O365, VG dedup and ImageNetBoxes as detection data. The KD source column shows the VLMs used by each method, in which the subscript `$_{text}$' means that the method used only the CLIP text encoder.}
\label{table:main_result_2}
\end{table*}

\subsection{The Effect of Pseudo Caption Labels}
\label{exp:main_1}

Table \ref{table:main_result_1} compares the evaluation results of the baseline and PCL models.
The PCL model trained on the LVIS base improves APr from 18.7 to 20.1 compared to the baseline model.
There is a decrease in AP from 34.5 to 33.5, which was anticipated since the model was exposed to manual prompt labels only 20\% of the time during training, which have the same patterns as those used during testing.

On the other hand, the PCL model trained on the de-duplicated VG dataset shows a significant improvement, with an increase in APr from 21.1 to 25.7.
It even outperforms the baseline model trained with all categories (Baseline\dag).
Although the baseline model was not heavily tuned, it is still an intriguing result.
We speculate that the PCL model was able better generalize to unseen object classes by leveraging more comprehensive knowledge extracted from VLMs using diverse pseudo caption labels generated from object instances over 80,000 categories.

Moreover, it is worth noting that the performance gap between the PCL models with and without de-duplication (25.7 vs. 28.5) suggests that there is still room for improvement in future research.

\subsection{Comparison with Other Methods}
\label{exp:main_2}

In this section, we provide a comprehensive comparison of PCL models with previous work, considering factors such as model sizes, detection datasets, image datasets, and sources of knowledge distillation.
The comparison targets for our model are previous works that have reported box-AP and box-APr, since our model does not produce segmentation results.
Table \ref{table:main_result_2} summarizes the comparison results, grouped by backbone sizes.
The fifth column denotes the detection datasets used to train the models.
O365, GoldG and OI stands for Object365~\cite{Object365_2019}, GoldG~\cite{Mdetr2021}, and OpenImages~\cite{OpenImages} datasets, respectively.
The sixth column shows the image-label or image-text pair datasets used as an additional source of information, including GLIP's 4 million image-text pair dataset (Cap4M)~\cite{GLIP2022}, a subset of ImageNet-21K filtered with LVIS categories (ImageNet-L)~\cite{Detic2022}, ConceptualCaptions' 3 million image-text pair dataset (CC3M)~\cite{CC3M2018}, ConceptualCaptions' 12 million image-text pair dataset (CC12M)~\cite{CC12M2021}, and Stony Brook University's 1 million photograph caption dataset (SBU)~\cite{SBU2011}, respectively.
The seventh column shows the image and text encoders used by each method.
GLIP~\cite{GLIP2022} does not use pre-trained encoders.
ViLD-text~\cite{ViLD2021}, VLDet~\cite{VLDet2023} and PCL methods use CLIP text encoders, while ViLD-ensemble~\cite{ViLD2021}, RegionCLIP~\cite{RegionCLIP2022}, OWL-ViT~\cite{OWLViT2022} use both CLIP image and text encoders.

Our best model achieved an APr of 30.6, as shown in the bottom row of Table \ref{table:main_result_2}, which is comparable to the state-of-the-art method of OWL-ViT (31.2).
This result indicates that by using PCL, our model was able to distill sufficient knowledge to attain the same level of strength as OWL-ViT.
Notably, we observed that the PCL model with a medium size backbone (Swin-B) performed only slightly worse than our best model (29.1 vs. 30.6), whereas OWL-ViT had a noticable performance gap (20.6 vs. 31.2).
This difference can be attributed to the fact that PCL extracts knowledge from the same large CLIP model (CLIP ViT-L/14 text encoder) regardless of the size of its backbone.

We also compared our best model with the PCL model trained on the LVIS base.
The performance gap (30.6 vs. 24.7) indicates that PCL requires a sufficiently large vocabulary data to have a significant impact.
Thus, a straightforward future research direction is to combine PCL with other OVD methods that expand the vocabulary size of data, such as RegionCLIP and VLDet.

\subsection{Ablation Studies}
\label{exp:ablations}

We conduct experiments to assess the effect of our design choices for PCL.
These include the number of labels assigned per object, the utilization of object class names as the prefix for the captioning model's output, the application of manual prompt labels, and the use of crop margin.
The experiments are conducted under the same conditions as the main experiment, using the de-duplicated VG dataset and the Swin-T backbone.

\begin{table}[t!]
\begin{center}
\begin{tabular}{c c c} 
 \hline
 Method & APr & AP \\
 \hline\hline
 PCL (ours)                              & 25.7 & 29.5 \\  
 20 $\Rightarrow$ 10 labels per object   & 24.3 (-1.4) & 29.0 (-0.5) \\
 20 $\Rightarrow$ 1 label per object     & 23.3 (-2.4) & 28.1 (-1.4) \\
 without output prefix                   & 23.6 (-2.1) & 26.3 (-3.2) \\
 without manual prompts                  & 23.7 (-2.0) & 27.3 (-2.2) \\
 20\% $\Rightarrow$ 0\% crop margin      & 24.1 (-1.6) & 29.1 (-0.4) \\
 \hline
\end{tabular}
\end{center}
\caption{Ablation study results. Second and third row shows the effect of the number of labels per object. The following rows describe the effect of using class name output prefix, manual prompts, and the crop margin.}
\label{table:ablation_results}
\end{table}

\subsubsection{The Impact of the Number of Labels}

Objects can be described from various viewpoints.
In the main experiment, we employed 20 labels per object, out of which 16 were pseudo caption labels, to account for this variability.

To investigate the effect of the number of labels per object, we compared the performance of PCL models trained with 1, 10 and 20 labels per object.
Specifically, for the PCL models utilizing 10 and 20 labels per object, we used 8 and 16 pseudo caption labels, respectively.
For the model with only 1 label per object, we assigned a pseudo caption label 80\% of the time and a manual prompt label 20\% of the time.

As shown in the second row of Table \ref{table:ablation_results}, the PCL model trained using 10 labels per object experienced a slight drop in APr from 25.7 to 24.3 (-1.4).
When the number of labels per object was reduced to 1, the performance dropped even more significantly from 25.7 to 23.3 (-2.4), as shown in the third row.
These results indicate that rich and diverse descriptions for an object are necessary to extract sufficient knowledge from VLMs.

\subsubsection{The Impact of Using Class Name as Output Prefix}
\label{exp:ablation_1}

\begin{figure}[t!]
    \centering
    \includegraphics[width=8cm]{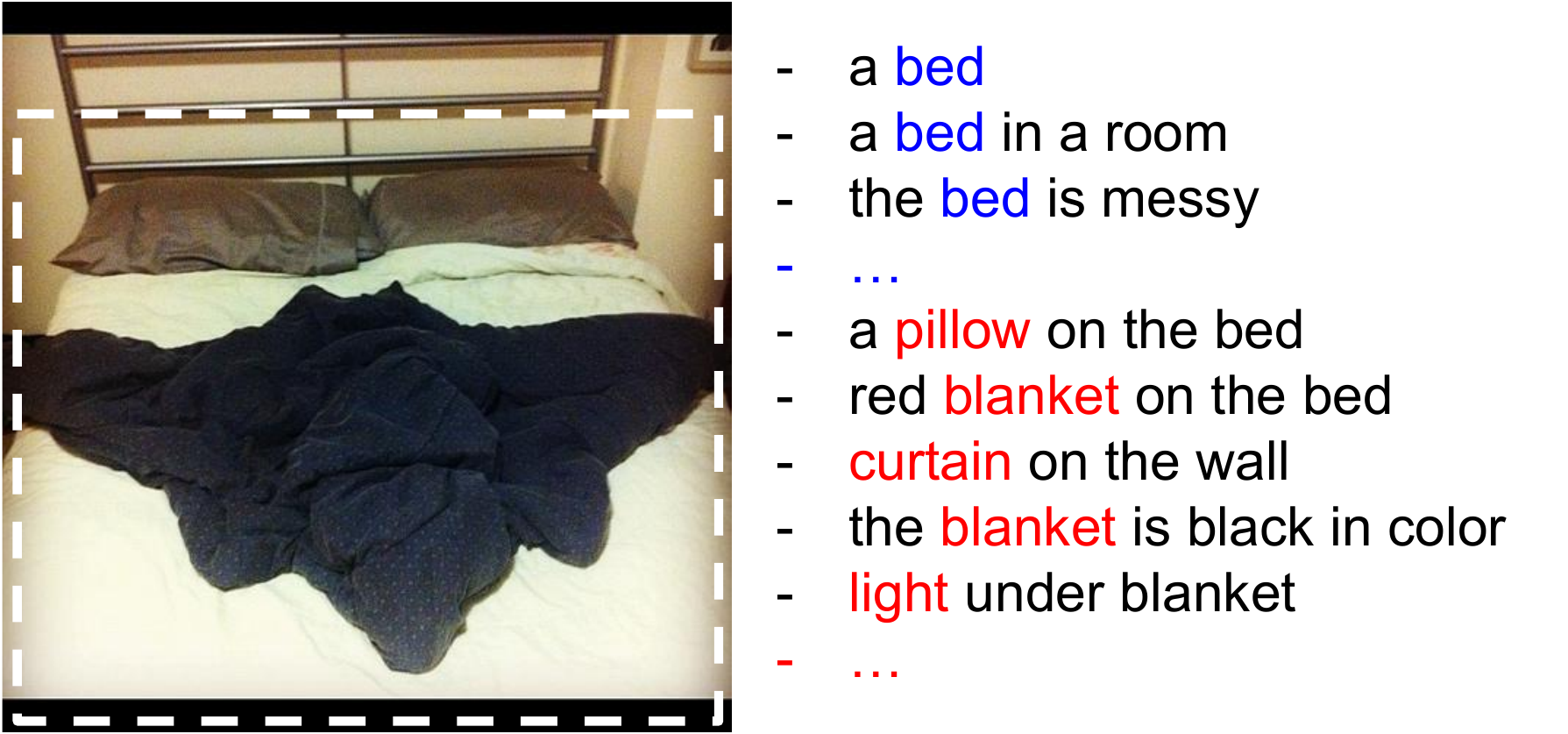}
    \caption{Pseudo caption labels for a photo of a \textbf{bed}. The captioning model generated inaccurate captions that referred to incorrect objects, such as pillows or blankets, since the output prefix did not include the object class name.}
    \label{fig:caption_labels_without_output_prefix}
\end{figure}

To generate pseudo caption labels, we imposed a constraint on the captioning model to output the object class name as the prefix.
This constraint is necessary since the model may generate captions that describe backgrounds or secondary objects located within the bounding box of the main object.
Figure \ref{fig:caption_labels_without_output_prefix} illustrates an example where captions describe secondary objects such as pillow, blanket, and curtain instead of the primary object, which is a bed.

We evaluated the impact of this constraint by training a PCL model with pseudo caption labels generated without the output prefix constraint.
The fourth row in Table \ref{table:ablation_results} shows that model performance significantly deteriorates in both AP (-3.2) and APr (-2.1) compared to the original PCL model in the first row.
This decline in performance is more pronounced in AP since many pseudo caption labels for known class objects now refer to incorrect object names.

\subsubsection{The Impact of Using Manual Prompt Labels}

To train a PCL model, we utilize a small portion of manual prompt labels in addition to the pseudo caption labels.
It is expected to reduce the discrepancy in target label patterns between the training and testing phases, resulting in improved detection performance for both seen and unseen object classes.

We can see the impact of using manual prompt labels by comparing two models in the first and fifth rows of Table \ref{table:ablation_results}.
The PCL model trained without manual prompt labels experienced significant performance drop in both AP (-2.2) and APr (-2.0).
This finding highlights the importance of maintaining label pattern consistency between training and testing for achieving high performance.

\subsubsection{The Impact of Image Crop Margin}

In addition to describing the characteristics of target objects, good pseudo caption labels also capture information about their surrounding environments.
An example is the label ``dog running on the grass'' in Figure~\ref{fig:system_overview}.
To allow the captioning model to better incorporate contextual information, we applied a 20\% margin to the object's bounding box area.

The impact of using the crop margin on the performance of the PCL model is shown in the last row of Table \ref{table:ablation_results}.
When we crop the object image using the original bounding box, APr experiences a significant drop of -1.6, while AP is less affected.
We believe that the model is able to recognize known category objects based on their characteristics, even when the context is not rich enough.

\subsection{Complex Freeform Text Detection}

\begin{figure*}
    \centering
    \includegraphics[width=0.9\textwidth]{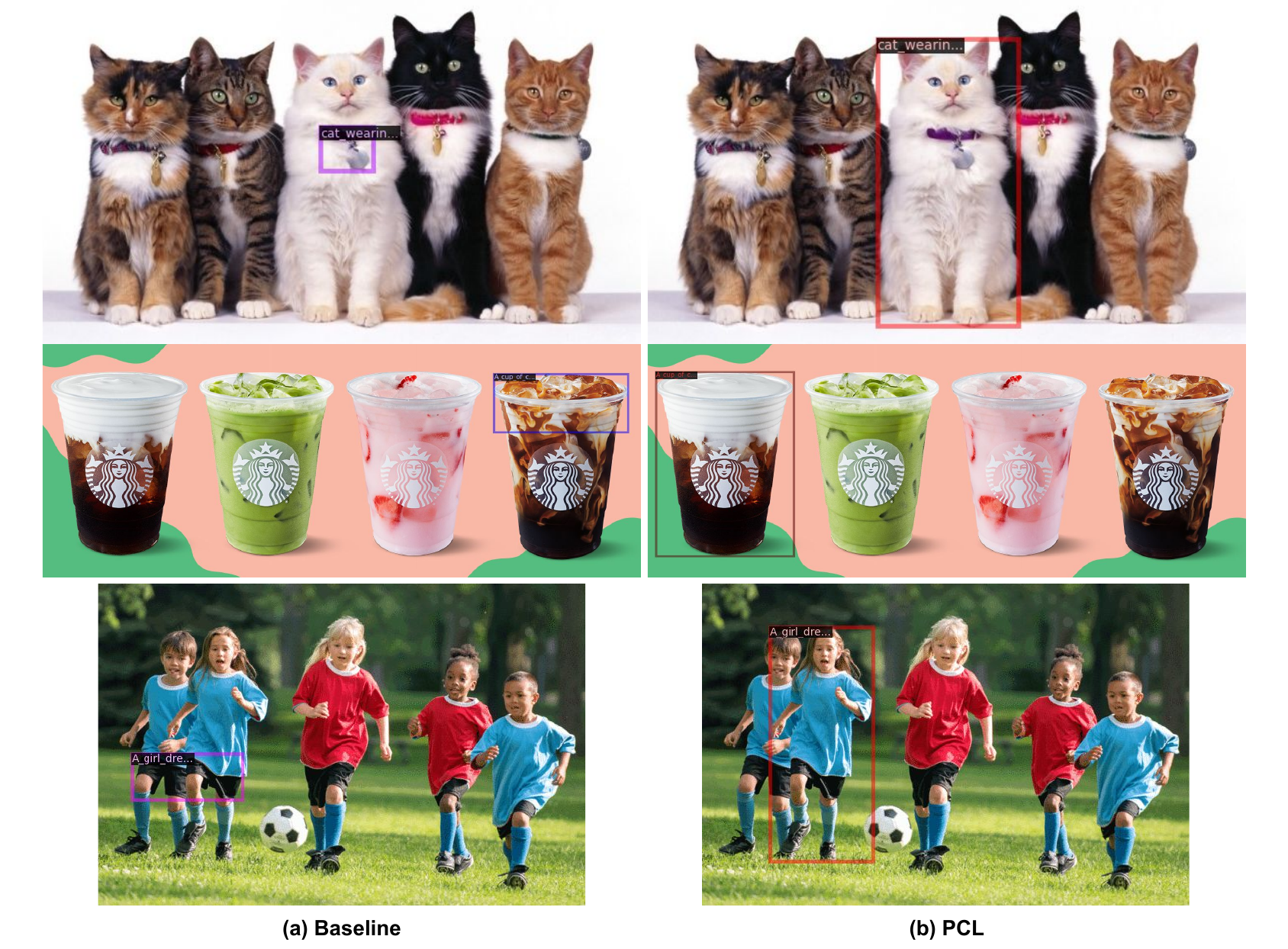}
    \caption{Comparison of the top-1 detection results of the baseline and PCL models with complex queries. The queries used were ``cat wearing a collar with a silver pendant'', ``a cup of coffee with a layer of white cream on top'', and ``a girl dressed in a blue shirt and black pants'', displayed from the top row. The comparison illustrates that the PCL model is capable of accurately distinguishing the main objects from other information in the queries.}
    \label{fig:freetext_detection}
\end{figure*}

The results of the LVIS benchmark indicate that PCL significantly enhances novel object detection performance.
This section provides a qualitative analysis demonstrating that PCL models understand the linguistic structure of an input query better than baseline models.

Figure \ref{fig:freetext_detection} compares the top-1 detection outcomes of the baseline and PCL models (with the Swin-T backbone).
The first row shows the results of the models with the query ``cat wearing a collar with a silver pendant''.
The models need to comprehend that the main object to detect is a ``cat'', and the ``collar'' and ``pendant'' are secondary objects that should be included. 
However, the baseline model mistakenly detects only ``a silver pendant'', while the PCL model accurately identifies the main object, which is a ``cat'' wearing ``a collar with a silver pendant''.
For the second and third samples, the same results are obtained when using queries ``a cup of coffee with a layer of white cream on top'' and ``a girl dressed in a blue shirt and black pants''.

This result demonstrates the ability of the PCL model to learn the linguistic structure from pseudo caption labels and use that knowledge to comprehend complex queries accurately.
This skill of PCL can potentially be employed for referring expression comprehension.

\section{Conclusion}
\label{sec:conclusion}

\begin{figure}[t!]
    \centering
    \includegraphics[width=8cm]{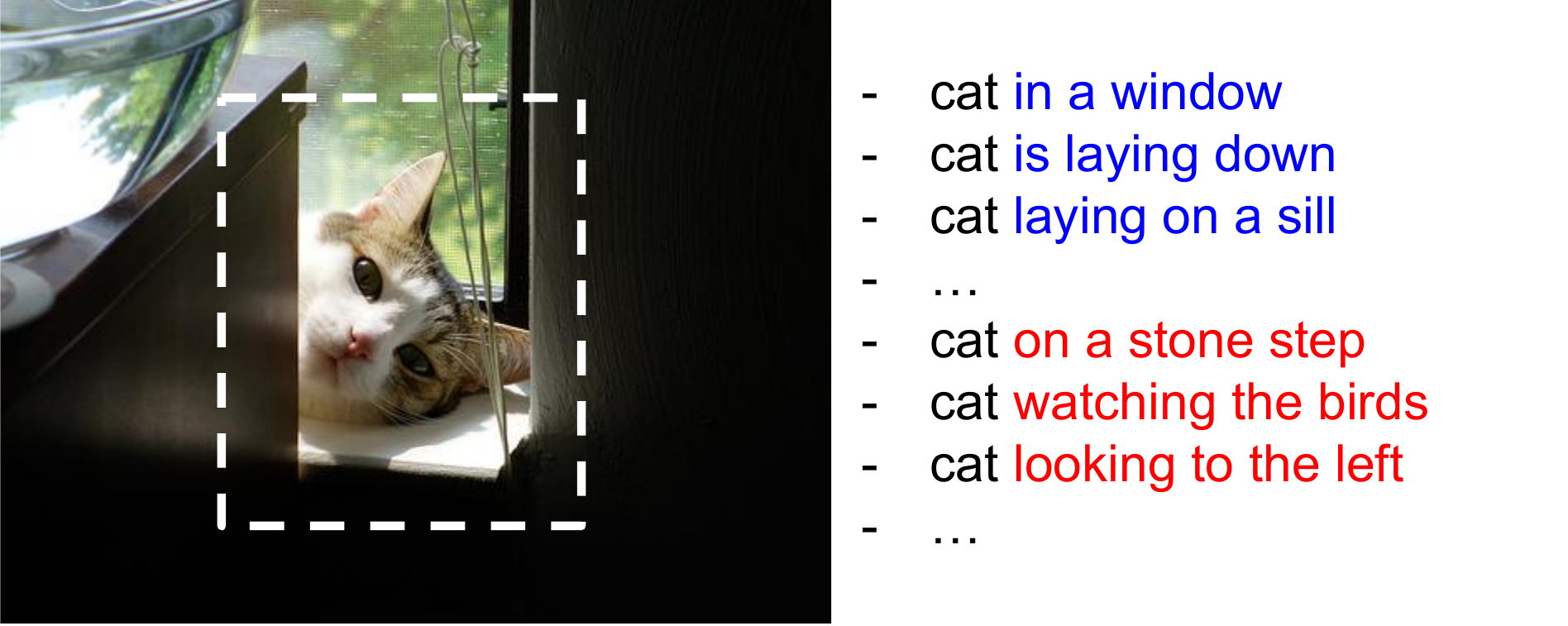}
    \caption{Pseudo caption labels for a photo of a \textbf{cat}. Depending on the captioning model's performance, generated captions may not correctly describe the object as shown in red color.}
    \label{fig:correct_and_incorrect_pseudo_captions}
\end{figure}

In this study, we presented a simple and effective knowledge distillation method, Pseudo Caption Labeling (PCL), for OVD.
PCL is a simple pre-processing technique that transforms a detection dataset to a region-caption dataset using an image-captioning model, making it possible to use well-established OVD model architectures and training recipes.
Our experiment results also demonstrated its effectiveness by achieving comparable performance to the state-of-the-art OVD methods.

While our proposed method has shown promising results, we also have identified areas that require further improvement.
Firstly, the captioning model often generates pseudo caption labels with inaccurate contextual information, as shown in Figure \ref{fig:correct_and_incorrect_pseudo_captions}.
For this specific example, we found that 5 were correct, 7 were incorrect and 4 were ambiguous among 16 pseudo caption labels.
This issue could potentially be addressed by using a more accurate and reliable captioning model.

Secondly, the output prefix constraint may restrict the syntactic diversity of the generated captions.
One potential solution is to sample a large number of captions without this constraint and then select those that contain the target class name as a sub-string.
Alternatively, we could develop a captioning model that explicitly takes object name as an input condition to generate more diverse captions while maintaining accuracy.

Lastly, a small portion of manual prompt labels were used during training to compensate for the discrepancy of target labels between training and testing phases.
A potential improvement to this approach is to use object-specific prompts at testing time utilizing large language models, as demonstrated in the previous work~\cite{VCwithLLM2023}.


\end{document}